\documentclass[letterpaper, 10 pt, conference]{ieeeconf}  

\IEEEoverridecommandlockouts                              

\usepackage{graphics} 
\usepackage{hyperref}
\usepackage{adjustbox}
\usepackage{caption}
\usepackage[compact]{titlesec} 
\usepackage{capt-of,etoolbox}
\makeatletter
\apptocmd\@maketitle{{\myfigure{}\par}}{}{}
\makeatother

\usepackage{caption}
\newcommand\myfigure{%
\centering
 \includegraphics[width=\textwidth]{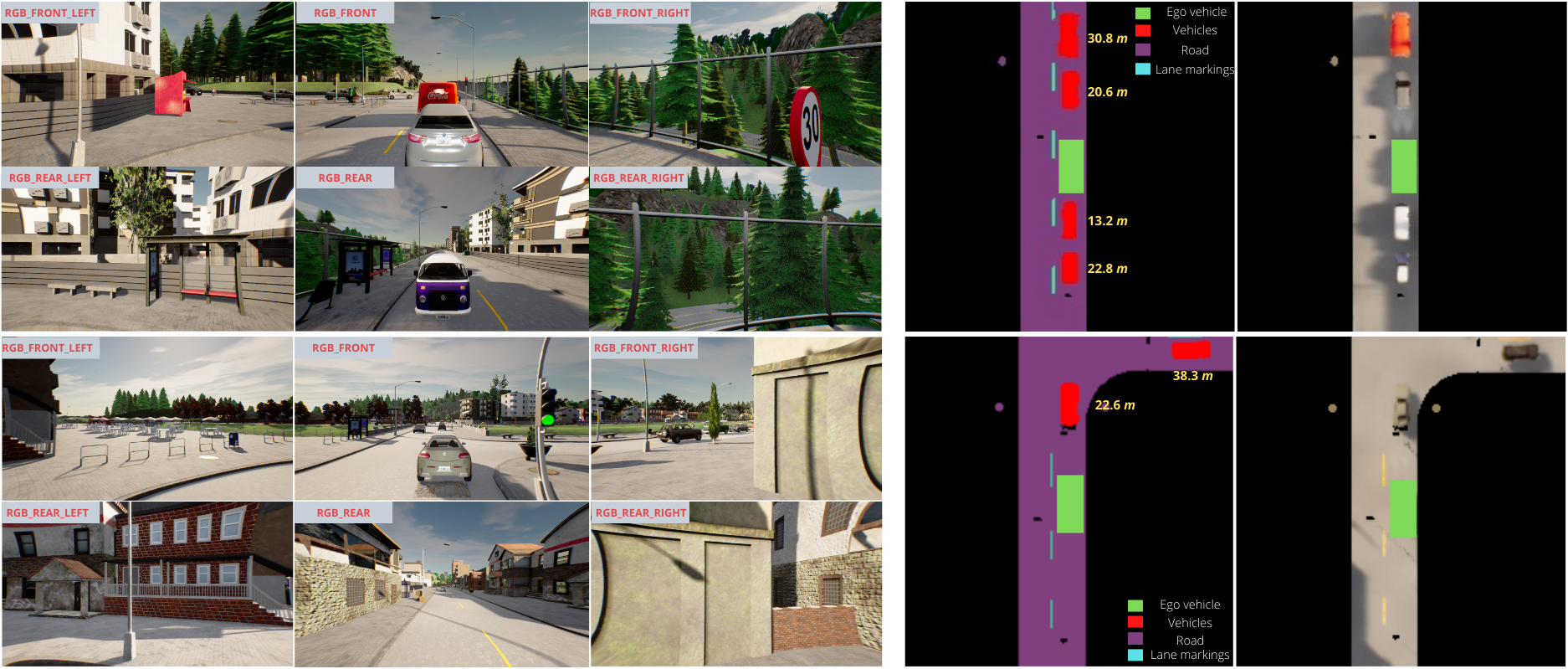}
  \captionof{figure}{\textbf{Appearance and occupancy information from monocular camera with surround FOV.}  We show qualitative results of our method, trained on Carla \cite{carla} dataset. \textit{Left} 
  The input images to our system are monocular RGB images ($6$ in our case - \textit{front left, front, front right, rear left, rear, and rear right}), recorded from cameras mounted on the top of the vehicle. \textit{Right} First, we show the dense semantic occupancy information predicted by our network, reasoning about \textit{vehicles, road, and lane markings}. Next, we show the appearance (color) information for the same occupancy grid. For every vehicle in the occupancy grid, we display the longitudinal distance of the centroid of a tight-bounding box around it from the ego frame origin. Best viewed digitally.}
  \label{fig:teaser}
}

\usepackage{amssymb}  
\usepackage{mathtools}
\usepackage{xcolor}
\usepackage{graphicx}
\usepackage{float}
\usepackage{gensymb}
\begin{document}

\title{\LARGE \bf
Estimation of Appearance and Occupancy Information in Bird's Eye View from Surround Monocular Images}
\author{Sarthak Sharma$^{1}$, Unnikrishnan R. Nair$^{1}$, Udit Singh Parihar$^{1}$, Midhun Menon S$^{1}$ and Srikanth Vidapanakal$^{1}$
\thanks{$^{1}$Authors are with the Ola Electric AI, Bengaluru,India.
{\tt\small \{sarthak.sharma1, unnikrishnan.r, udit.parihar, midhun.s, srikanth.vidapanakal\}@olaelectric.com. }}%
}

\maketitle

\begin{abstract}

Autonomous driving requires efficient reasoning about the location and appearance of the different agents in the scene, which aids in downstream tasks such as object detection, object tracking, and path planning. The past few years have witnessed a surge in approaches that combine the different task-based modules of the classic self-driving stack into an End-to-End(E2E) trainable learning system. These approaches replace perception, prediction, and sensor fusion modules with a single contiguous module with shared latent space embedding, from which one extracts a human-interpretable representation of the scene. One of the most popular representations is the Bird's-eye View (BEV), which expresses the location of different traffic participants in the ego vehicle frame from a top-down view. However, a BEV does not capture the chromatic appearance information of the participants. 

To overcome this limitation, we propose a novel representation that captures various traffic participants' appearance and occupancy information from an array of monocular cameras covering $360^\circ$ field of view (FOV). We use a learned image embedding of all camera images to generate a BEV of the scene at any instant that captures both appearance and occupancy of the scene, which can aid in downstream tasks such as object tracking and executing language-based commands.

We test the efficacy of our approach on synthetic dataset generated from CARLA. The code, data set, and results can be found at \href{https://rebrand.ly/APP_OCC-results}{https://rebrand.ly/APP\_OCC-results}.

\end{abstract}


\section{Introduction}
Autonomous driving is one of the most active areas of research. The software stack for autonomous driving has evolved a lot since the first DARPA Grand Challenge 2004. From a modular architecture comprising a cascade of task-specific blocks (sensor fusion, perception, planning, and control), it has evolved to an E2E system that learns to generate driving behaviors from perceived sensor inputs. E2E systems can produce diverse and complex driving behaviors because it learns by mimicking expert behaviors. However, driving is a highly evolved and context-sensitive task, and hence, a direct regression over behaviors does not converge quickly or easily. Researchers have successfully used shared embeddings with trained auxiliary interpretable hierarchical tasks to overcome this problem. One of such promising interpretable intermediate representations is BEV\cite{faf, lss, fiery, neat, nmr, urtasunmp3, urtasunp3}. 
\setcounter{figure}{1}
\begin{figure*}[t!]
  \includegraphics[width=\textwidth,height=8.2cm]{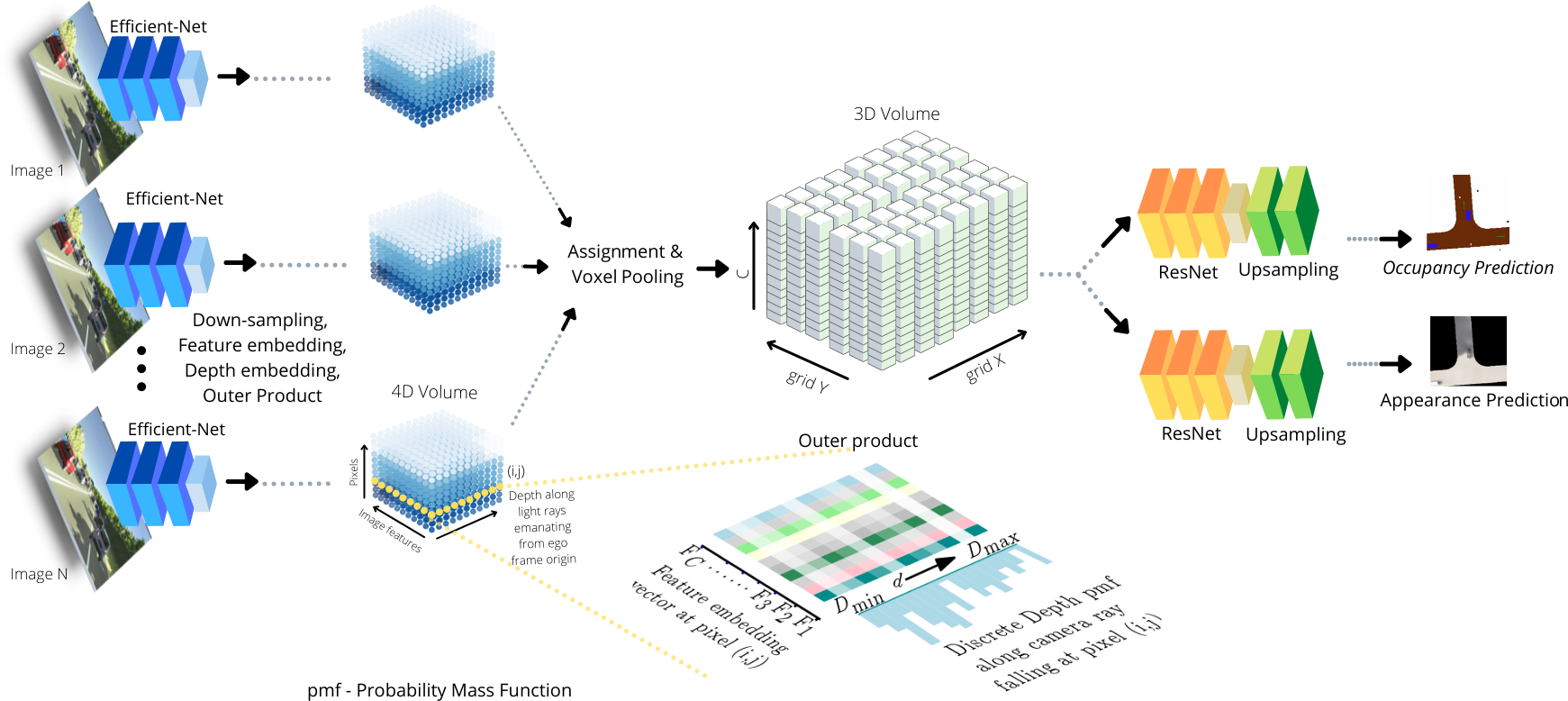}
  \captionsetup{singlelinecheck=off}
  \caption{ \textbf{Architecture}: Proposed architecture for estimating color and occupancy from monocular camera inputs.}
  \begin{itemize}
  \item We pass each camera image $I_t^i\in \mathbb{R}^{H \times W \times 3}$ through an  Efficient-net-B0\cite{efficientnet} backbone to obtain feature embedding $\varepsilon_t^i$. 
  \item We then split these camera features into a context vector $^{\text{context}}\varepsilon_t^i$ and a depth distribution $^{\text{depth}}\varepsilon_t^i$.
  \item We derive a self-attended discrete depth weighed features on the $1:16$ image using the outer product $^{\text{context}}\varepsilon_t^i\otimes^{\text{depth}}\varepsilon_t^i$.
  \item Once all the N cameras have been processed, we perform voxel pooling at pre-defined grid resolution $\text{grid}_X \times \text{grid}_Y$ to obtain a tensor of dimension $C \times \text{grid}_X \times \text{grid}_Y$. 
  \item We finally output Occupancy $\mathcal{O}$ and appearance $\mathcal{A}$ using a Res-Net\cite{resnet} backbone, followed by up-sampling layers. Cross-Entropy loss supervises the occupancy prediction and $L_1$ loss supervises the appearance prediction.
  \end{itemize}
  \label{fig:model}
\end{figure*}
BEV is a top-down orthographic view of the space around the Self-Driving Vehicle (SDV) in an egocentric frame of reference. It is averse to occlusion effects (a significant drawback of perspective projection), and being a representation in metric space, it is amenable to planning. The existing approaches in BEV \cite{neat,lss,fiery, nmr} aim to capture the occupancy information of different traffic participants in the scene, such as vehicles, roads, and lane markings. While this information provides semantic occupancy and shape understanding (in the form of object dimensions, orientation), it lacks appearance information. 

Appearance and location are vital cues that humans infer from the environment. We use them to reason about the environment and its different actors heavily. We use them to label/tag the actors and track their temporal behavior. 
With steeringless and pedalless cars around the corner, \cite{news1, news2, admin.2022}, it is only natural that Multi-Object Tracking (MOT)\cite{gridtrack,real-time-bev-track,beyondpixels} and language-based navigation \cite{2020commands,2020giving,talk2car,unni_1,unni_2} will be inevitable features for any future SDV. Complementary appearance cues present in the BEV occupancy space and the RGB image space provide strong priors for the above-mentioned tasks. 

In MOT, existing work focuses mainly on costs derived from occupancy information such as position, shape, and orientation \cite{gridtrack,real-time-bev-track,beyondpixels}. In \cite{beyondpixels}, to exploit the appearance cues, the authors back-project different object instances in BEV into image space. This can lead to significant overlap of regions and weak correlation during occluded object scenarios. 

For tasks such as language-based navigation \cite{2020commands,2020giving,talk2car,unni_1,unni_2} (e.g. \textit{You can park up ahead behind the \textbf{silver car}, next to the lamp post with the orange sign on it}), appearance and semantic knowledge are critical to create associations. The image space captures appearance information, but it lacks metric understanding of the scene due to perspective distortion. BEV captures the semantic information in metric space in the form of an occupancy grid but lacks appearance priors.

In this work, we present, to the best of our knowledge, a novel BEV representation that captures the appearance and occupancy information of the traffic participants.
Focusing on the relevant classes of traffic participants $S \in$ \{road, vehicles and lane markings\}, we present a representation that can densely reason about the appearance and color of these participants from multiple monocular cameras that cover a $360^{\circ}$ FOV. The occupancy information determines the probability of belonging to the classes in $S$, for each cell of the grid. We learn to capture the information about the chromatic appearance as RGB triplets for the classes in $S$.

Our core contributions are as follows.
\begin{itemize}
  \item We propose, to the best of our knowledge, the first such unified dense BEV representation that captures the appearance and occupancy of different traffic participants in the scene. The proposed architecture injests images from monocular cameras with overlapping FOV and the respective camera parameters (extrinsic and intrinsic).
  \item We demonstrate the positive coupling between the occupancy and appearance training tasks. Using ablation studies, we quantitatively show how the capture of the image information of the scene improves the estimate of the occupancy information.
    \item We test our method on a synthetic data set generated from Carla \cite{carla} and report qualitative and quantitative results for occupancy estimation. We compare our method with two SOTA approaches \cite{neat,lss} and report a significant performance improvement.
    \item We propose a pipeline for generating training data for the proposed architecture from the NuScenes dataset. \cite{nuscenes} and show qualitative results for the same.
\end{itemize}

\section{Related Work}
\subsection{Occupancy information in BEV space}
There has been a recent surge in approaches that reason about the occupancy in a BEV space using images from a monocular camera array with combined surround FOV. In \cite{neat}, the architecture learns an implicit representation of the scene semantics in BEV. The network trains on the data captured from an array of monocular cameras facing the front of the SDV with a combined 180$^\circ$ FOV. They assist learning by using an auxiliary task of waypoint prediction. In approaches such as \cite{lss,fiery}, the authors do dense semantic reasoning in the BEV grid space using images from similar monocular camera rigs. The networks in \cite{urtasunp3,urtasunmp3} learn to predict the spatiotemporal dynamics of the scene on a BEV grid using LiDAR point clouds as input. However, these approaches do not capture the color information of the different traffic participants.

\subsection{Appearance information in BEV space}
 In \cite{zisserman}, the authors propose a method to obtain the BEV of a scene from a single perspective image. A Convolutional Neural Network (CNN) estimates the image's vertical vanishing point and the ground plane vanishing line (horizon). The vanishing point and the vanishing line of the ground plane determine the homography matrix $H$ that maps the image to the overhead view after removing the perspective distortion. However, the approach depends on the FOV of the camera. Furthermore, it does not reason about the semantic classes of different agents in the top-down view.
 
\begin{figure}
 \centering 
  \includegraphics[width=\columnwidth]{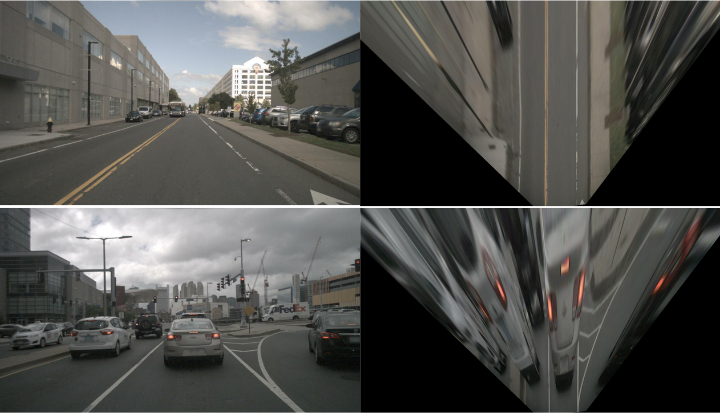}
  \caption{\textbf{Estimating BEV using IPM}: We show results of obtaining BEV using IPM on NuScenes \cite{nuscenes}, where we calculate the homography matrix using the planar road points. \textit{Top} When the world is mainly planar around the ego-vehicle, the BEV estimate of the scene captures the information well. \textit{Bottom} Once the planarity assumption of the road gets violated, the BEV estimate accuracy drops drastically.}
  \label{fig:ipm}
\end{figure}

 Classical methods like Inverse Perspective Mapping (IPM) \cite{mvg} reason about the BEV of the scene by estimating the homography matrix based on the road surface. While this method leads to a plausible BEV of the planar road surface(Fig \ref{fig:ipm}, \textit{Top}), it heavily distorts the other traffic participants, as shown in Fig \ref{fig:ipm}. The semantic information of the scene is also not available.

\section{Approach}\label{approach}
\subsection{Synthetic data generation}

\begin{figure}
 \centering 
  \includegraphics[width=\columnwidth]{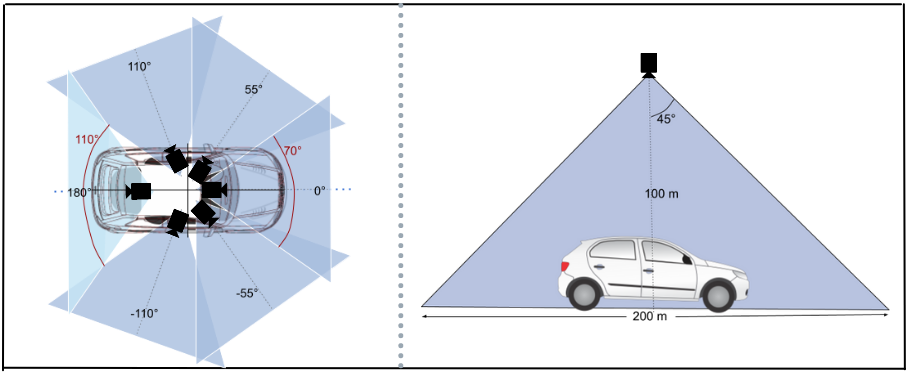}
  \caption{\textbf{Sensor setup for data collection for appearance and occupancy reasoning}: \textit{Left} We place $6$ RGB cameras on the ego vehicle, covering full $360^{\circ}$ surroundings. Out of the $6$ cameras, $5$ (\textit{front-left, front, front-right, rear-left and rear-right}) are of $70^{\circ}$ FOV, and $1$ (\textit{rear}) is of $110^{\circ}$ FOV. We show the FOV in red and the orientations of the camera in black. \textit{Right} We show the placement of our BEV camera, at the height of $100m$ above the ground, with a FOV of $90^{\circ}$. The BEV camera provides us with information regarding the appearance and occupancy of the scene.}
  \label{fig:sensor}
\end{figure}

We,
\begin{itemize}
    \item Collect monocular RGB camera images using an expert driving agent in CARLA \cite{carla}. We collect driving data on routes from the eight publicly available virtual towns for training and test data generation, randomly spawning scenarios at several locations along each route.
    \item Replicate the NuScenes \cite{nuscenes} data set for the intrinsic and extrinsic parameters of the camera. We place six cameras - five of them \{\textit{front-left, front, front-right, rear-left and rear-right}\} with a FOV of 70$^{\circ}$ and one (\textit{rear}) with an FOV of 110$^{\circ}$. The camera orientations are $55^{\circ},0^{\circ},-55^{\circ},110^{\circ},-110^{\circ},180^{\circ}$ respectively. We show the sensor setup in Fig. \ref{fig:sensor}. Images have a resolution of $1920 \times 1080$. 
    \item To capture the color BEV of the scene, we place a camera at the height of $100$ meters above the car, pitching down at $90^\circ$. This BEV camera has a FOV of $90$ $^{\circ}$ and captures top-down RGB and semantic views. We capture these images at a resolution of $1000 \times 1000$ pixels, focusing on an area of $200m \times 200m$ around the vehicle. We use a $400 \times 400$ center-crop  resized to a $200 \times 200$ resolution image, finally capturing an area $80m \times 80m$ around the vehicle. 
\end{itemize}

\subsection{Model Architecture}
Motivated by [3], our architecture infers the occupancy and appearance information from multiple monocular images $180^\circ$ FOV. Inputs to our model are the camera images and camera parameters (extrinsic and intrinsic). The model then deduces the appearance and occupancy of the scene in BEV. We show the model architecture in Figure \ref{fig:model}.

\textbf{Encoding image features}: At the current time $t$, we have RGB images $I=\left\{I_t^i| i=\{1...N\}\right\}$, where N=number of cameras, and $I_t^i\in \mathbb{R}^{H \times W \times 3}$ ($W$ and $H$ are the width and height of each image). To extract features, we pass each image $I_t^i$ through a standard convolution encoder $E$ (using Efficient-net-B0 \cite{efficientnet}). The encoder then down-samples the input by a factor of $16$ and generates a feature embedding $^{\text{context}}\varepsilon_t^i$ = $E(I_t^i) \in \mathbb{R}^{C \times H' \times W'}$ where $\displaystyle{H'=\frac{H}{16}}$ and $\displaystyle{W'=\frac{W}{16}}$ and $C$ = number of image features. We also reason about a discrete probability distribution of metric depth $d\in[D_{\text{min}},D_{\text{max}}]$ of each projected pixel (to a resolution of $\Delta d$) on the down sampled image. We denote the depth embedding  as $^{\text{depth}}\varepsilon_t^i \in \mathbb{R}^{D \times H' \times W'}$ where $\displaystyle{D=\frac{D_{\text{max}}-D_{\text{min}}}{\Delta d}}$. At each pixel index $(i,j)$ in the $H' \times W'$ downsampled image, there exists a context vector $c_{ij} \in \mathbb{R}^{C} $ and corresponding discrete depth distribution $d_{ij} \in \mathbb{R}^{D}$. The network first learns the stacked tensor embedding $\varepsilon_t^i = \left[^{\text{context}}\varepsilon_t^i,~^{\text{depth}}\varepsilon_t^i\right]$. Next, using the outer product $\gamma_t^i=^{\text{context}}\varepsilon_t^i\otimes^{\text{depth}}\varepsilon_t^j\in \mathbb{R}^{C \times D \times H' \times W'}$, we capture the camera features modulated by the discrete depth probabilities to form an approximation for self-attention for context features at each down-sampled image pixel.
For each camera, we lift the set of outer product vectors $\left\{\gamma _1,\gamma_2,\dots,\gamma_N\right\}$ of the weighed features for all image pixels using the known intrinsic and extrinsic parameters of the camera to bring them to the ego-frame. 

\begin{figure*}[ht!]
  \includegraphics[width=\textwidth]{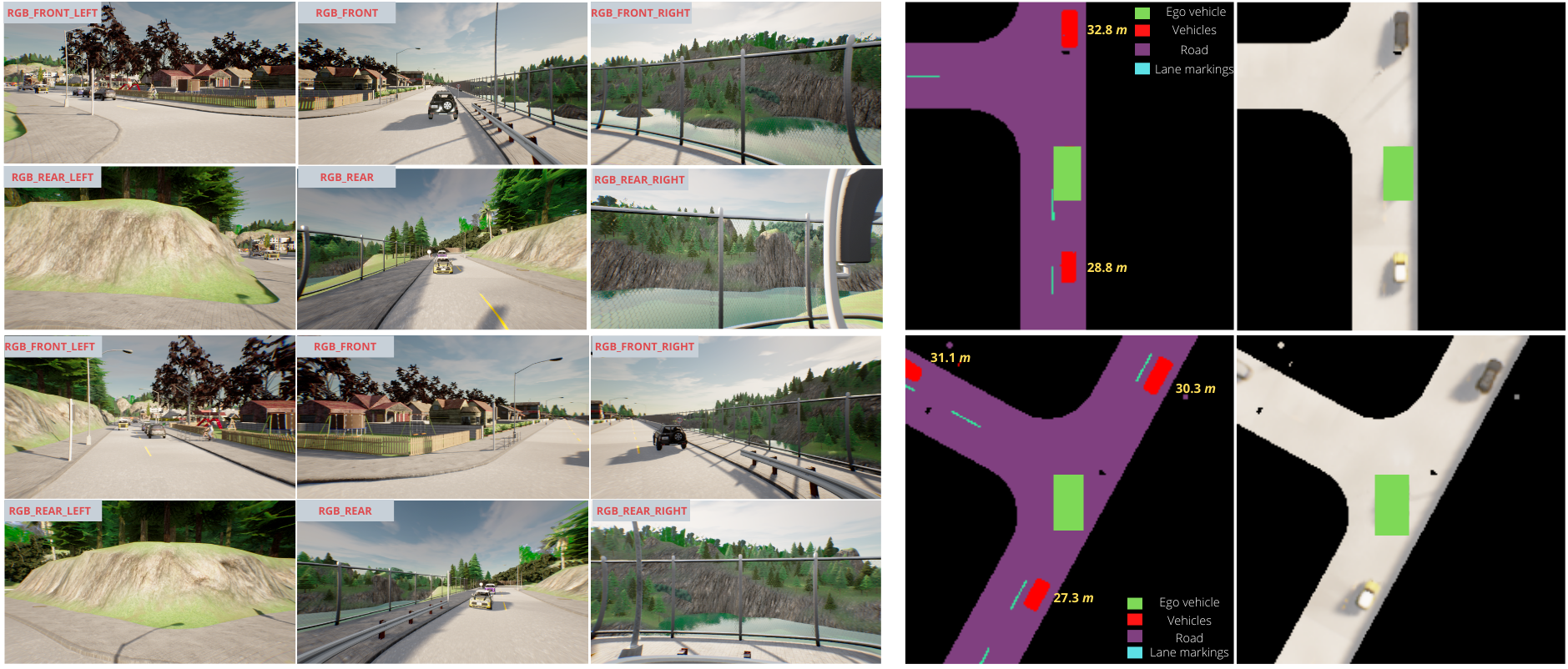}
  \caption{\textbf{Qualitative results of our method, trained on the Carla [1] dataset}: We could capture the BEV image with occupancy and complete appearance information for different turning and intersection scenarios. We display the longitudinal distance of the vehicle centroid in the occupancy grid from the center of the ego vehicle. Best viewed digitally. }
  \label{fig:qual2}
\end{figure*}
\textbf{BEV projection}: 
We transform the lifted set of features $\left\{\gamma _1,\gamma_2,\dots,\gamma_N\right\}$ into the BEV. We define the extent of the BEV grid as $\epsilon_{X} \times \epsilon_{Y}$, centered on the ego vehicle. We discretize this area into a unit grid of resolution $\nabla_{X} \times \nabla_{Y}$, resulting in a grid of dimension $\text{grid}_X \times \text{grid}_Y$. The above point cloud obtained from the camera encoder is then Voxel pooled, as described in \cite{pointpillars} to obtain a tensor of dimension $C \times \text{grid}_X \times \text{grid}_Y$. 

\textbf{Appearance and occupancy information}: 
From the above-pooled tensor, we extract the appearance and occupancy information. To obtain the occupancy tensor $\mathcal{O}$ with dimension $ |S| \times \text{grid}_X \times \text{grid}_Y$(where $|.|$ indicates cardinality), we pass the above tensor of dimensions $C \times \text{grid}_X \times \text{grid}_Y$ through the first three layers of the ResNet-18 \cite{resnet} backbone, followed by two up-sampling layers.
We follow the same paradigm for the appearance reasoning tensor $\mathcal{A}$ - passing the projected BEV tensor through the first three layers of ResNet-18 [14], followed by two up-sampling layers to get $n_c \times \text{grid}_X \times \text{grid}_Y$ tensor (where $n_c$ = \#color channels). 

While the appearance tensor $\mathcal{A}$ is penalized for $L_1$ loss, the occupancy tensor $\mathcal{O}$ is penalized for cross-entropy (CE) loss.

\section{Experiments and Results}

We carried out extensive experiments of our approach on data collected from Carla \cite{carla}. We record $30$ sequences comprising over $12K$ scene instances. A scene instance comprises a set of N RGB images at any instant with $360^\circ$ FOV, along with their corresponding occupancy and appearance information from the BEV camera image. We make a $80-20$ split for training and test data. For our experiments, parameters we chose are $N=6,H=128, W=352, C=64, \epsilon_{X} = 80m, \epsilon_{Y}=80m, \nabla_{X}=0.4, \nabla_{Y}=0.4, grid_X = 200, grid_Y=200$. We sample depth values $d$ in the range $D_{min}=4$ and $D_{max}=45$, spaced $1.0 m$ apart. 
We train our method, as described in Section \ref{approach}, using PyTorch \cite{pytorch} framework. We use Adam optimizer \cite{adam} with a learning rate of $1e-3$ and a batch size of $20$.

\subsection{Appearance and Occupancy information}
We show quantitative and qualitative results of our approach in the following section. 

\begin{table}
    \centering
    \begin{adjustbox}{min width=1\linewidth}
    \label{multiprogram}
    \begin{tabular}{c|c|c|c|c|}
        \cline{2-4}
         & \multicolumn{3}{|c|}{Intersection-over-Union (IoU)}\\
        \cline{2-4}
         & Road & Vehicle & Lane  \\
        \hline
        \multicolumn{1}{|c|}{Lift-Splat-Shoot \cite{lss}} & $83.4$ & $37.67$ & $11.29$  \\
        \hline
        \multicolumn{1}{|c|}{\textit{Ours}} & $\mathbf{86.77}$ & $\mathbf{38.99}$ & $\mathbf{18.05}$ \\
        
        \hline
    \end{tabular}
    \end{adjustbox}
    \captionsetup{justification=centering}
     \caption{BEV occupancy prediction on Carla \cite{carla} in full surround $360\degree$ ($6$ cameras setting). 
    }
     \label{tab:iou-full}
\end{table}

In Table \ref{tab:iou-full} and Table \ref{tab:iou-half}, we show the quantitative numbers for the Intersection over Union (IoU) metric for the prediction of occupancy on our test split. We compare our prediction of BEV occupancy with \cite{lss,neat} using Intersection-over-Union (IoU) as the metric. Lift-Splat-Shoot\cite{lss} reports the metrics on NuScenes \cite{nuscenes}, hence we re-train the method of \cite{lss} for occupancy prediction for classes $S \in$ \{road, vehicles and lane markings\} on our data collected from Carla \cite{carla}. For NEAT\cite{neat}, the authors provide a pre-trained model trained on Carla \cite{carla} based dataset. 
\begin{table}
    \centering
    \begin{adjustbox}{min width=1\linewidth}
    \label{multiprogram}
    \begin{tabular}{c|c|c|c|c|}
        \cline{2-4}
         & \multicolumn{3}{|c|}{Intersection-over-Union (IoU)}\\
        \cline{2-4}
         & Road & Vehicle & Lane  \\
        \hline
        \multicolumn{1}{|c|}{NEAT \cite{neat}} & $58.1$ & $\mathbf{76.3}$ & $--$  \\
        \hline
        \multicolumn{1}{|c|}{Lift-Splat-Shoot \cite{lss}} & $63.53$ & $28.57$ & $12.27$  \\
        \hline
        \multicolumn{1}{|c|}{\textit{Ours}} & $\mathbf{68.48}$ & $32.49$ & $\mathbf{17.23}$ \\
        
        \hline
    \end{tabular}
    \end{adjustbox}
    \captionsetup{justification=centering}
     \caption{BEV occupancy prediction on Carla \cite{carla} for  $180\degree$ towards the front of the ego-vehicle ($3$ cameras setting). 
     }
     \label{tab:iou-half}
\end{table}

In Table \ref{tab:iou-full}, we compare our method with \cite{lss}. The approaches listed in Table \ref{tab:iou-full} take as input $6$ monocular camera images (along with their extrinsic and intrinsic parameters), covering full surround $360\degree$ of the ego vehicle and reason about the information in an $80m\times80m$ grid around the ego vehicle. We train our method for both appearance and occupancy prediction, whereas we train \cite{lss} for occupancy prediction. We report better performance for all the classes $S \in$ \{road, vehicles and lane markings\}, improving by $3.37, 1.32$ and $6.76$ respectively.

\begin{figure}
 \centering 
  \includegraphics[width=\columnwidth]{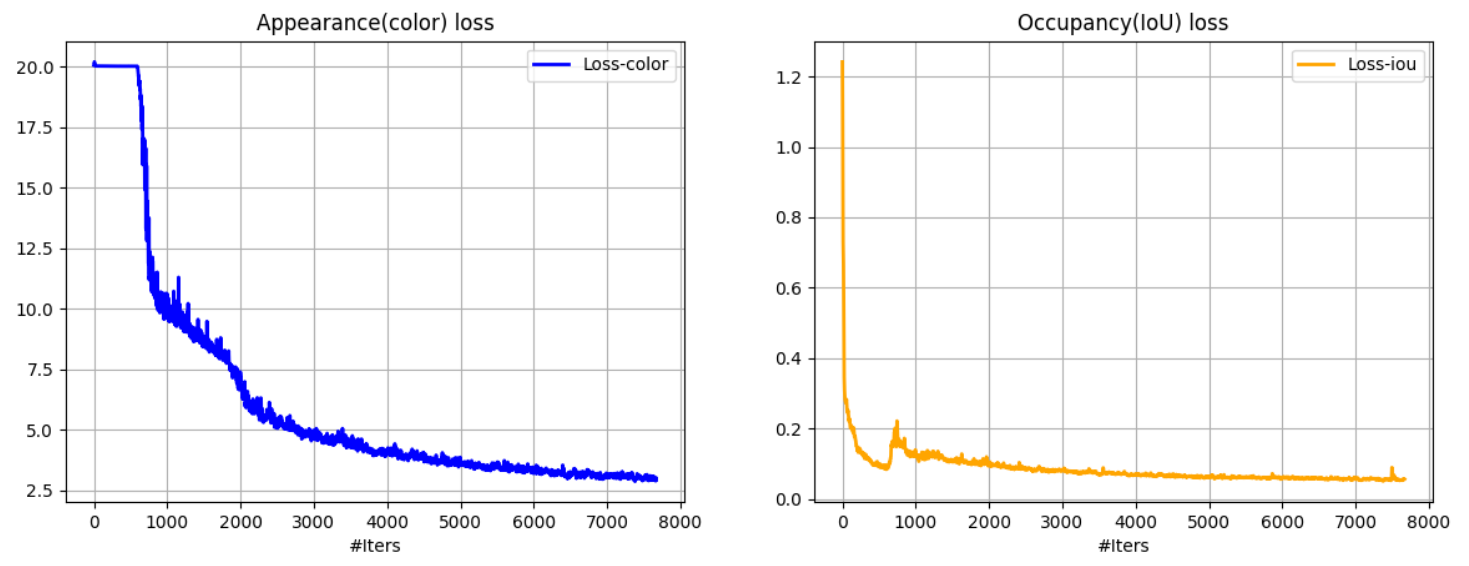}
  \caption{\textbf{Quantitative Result for Appearance and Occupancy Loss}: Here, we show the loss values for appearance (color) and occupancy (IoU) for our val split for surround $360\degree$ setting. The occupancy loss decreases from an initial value of $1.2$ to $0.05$ after 7K iterations. The appearance loss decreases from an initial value of $20.12$ to $3.046$ after the same number of iterations.}
  \label{fig:quan_loss}
\end{figure}

In Table \ref{tab:iou-half}, we compare our method with \cite{neat} and \cite{lss}. For \cite{neat}, the authors provide a pre-trained model trained on Carla \cite{carla}, which simultaneously reasons about occupancy and waypoint prediction for roads and vehicles. As the FOV for observation is $180^\circ$ front-facing, we re-train the method of \cite{lss} to take as input $3$ monocular camera images (along with their extrinsic and intrinsic parameters), covering front surround $180\degree$ of the ego vehicle and reason about the information in a $40m\times80m$ ego centric grid. We train our method in a similar setting, reasoning about appearance and occupancy information. Compared with \cite{lss}, our model exhibits a better performance for all the classes $S \in$ \{road, vehicles and lane markings\}, by $4.95$, $3.92$ and $4.96$ respectively. Compared with \cite{neat}, our performance improves by $10.38$ on the road, but our performance degrades by $43.81$ for the vehicle class. NEAT \cite{neat} does not report results on the class of lanes. We get an appearance loss of $3.046$ for the $360\degree$ surround setting and $2.89$ for the front $180\degree$ setting. We show the loss plot in Fig \ref{fig:quan_loss}.

In Fig. \ref{fig:teaser} and Fig. \ref{fig:qual2}, we show the qualitative results of our method in different traffic and intersection scenarios. We could capture appearance and occupancy information on straight roads, intersections, and turn scenarios. For each vehicle in the occupancy grid, we also display the estimated longitudinal distance of its centroid, computed from the center of the ego vehicle (shown in green).

\begin{table}
    \centering
    \begin{adjustbox}{min width=1\linewidth}
    \label{multiprogram}
    \begin{tabular}{c|c|c|c|c|}
        \cline{2-4}
         & \multicolumn{3}{|c|}{Intersection-over-Union (IoU)}\\
        \cline{2-4}
         & Road & Vehicle & Lane  \\
        \hline
        \multicolumn{1}{|c|}{occupancy ($180\degree$)} & $63.53$ & $28.57$ & $12.27$  \\
        
        \multicolumn{1}{|c|}{occupancy + appearance ($180\degree$)} & $\mathbf{68.48}$ & $\mathbf{32.49}$ & $\mathbf{17.23}$  \\
        \hline
        \hline
        \multicolumn{1}{|c|}{occupancy ($360\degree$)} & $83.4$ & $37.67$ & $11.29$ \\ 
        \multicolumn{1}{|c|}{occupancy + appearance ($360\degree$)} & $\mathbf{86.77}$ & $\mathbf{38.99}$ & $\mathbf{18.05}$ \\
        
        \hline
    \end{tabular}
    \end{adjustbox}
    \captionsetup{justification=centering}
     \caption{Ablation study. Comparing reasoning about occupancy and appearance information in \textit{(a)}. $180\degree$ front-facing   ($3$ cameras setting) \textit{(b)}. $360\degree$ surround ($6$ cameras setting).
     }
     \label{tab:abl}
\end{table}

\subsection{Ablation Studies}
We perform an ablation study of our network trained on two camera configurations to validate how reasoning about appearance aids in reasoning about occupancy. In configuration 1, we reason in front of the ego vehicle (covering $180\degree$ FOV). In configuration 2, we reason in a surround setting around the ego-vehicle (covering $360\degree$). We summarize the results in Table \ref{tab:abl}. For both the settings, we train two models-one that argues only about the occupancy and the other that argues about both occupancy and appearance. We consistently observe that incorporating appearance improves the reasoning about the occupancy of all classes $S \in$ \{road, vehicles, and lane markings\} in both the settings. 
\begin{figure*}[ht!]
  \includegraphics[width=\textwidth]{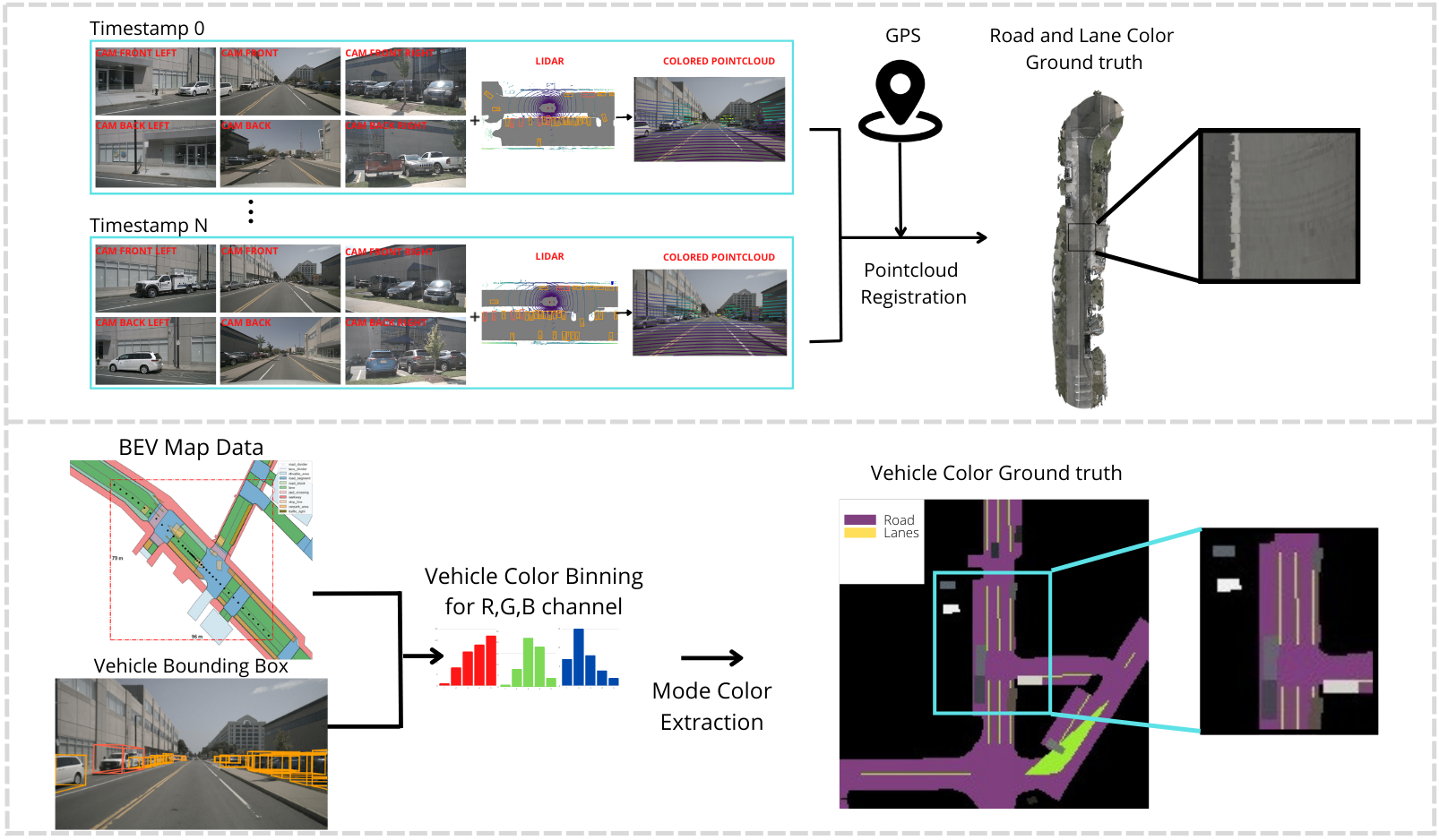}
  \caption{\textbf{Pipeline for generating data from a real-world dataset, for, e.g., NuScenes \cite{nuscenes} to capture appearance and occupancy information in BEV}: \textit{Top} We colorize LiDAR point clouds by projecting them into the camera images assigning each point the nearest pixel color. \textit{Bottom} We project the available annotated vehicle cuboids in the camera images, and for each of the $(R, G, B)$ channels, assign the post binning mode value. We could closely capture the colors of different vehicles and the detailed appearance of roads and lanes.}
  \label{fig:datagen}
\end{figure*}

\subsection{Real-world data generation}
For real-world datasets like \cite{nuscenes}, we split the generation of appearance information for BEV into two stages. For static classes of interest (like \textit{road, lanes}), we colorize the LiDAR point clouds by projecting them into the time-synced $6$ camera images and assigning every $(x, y, z)$ point in the point cloud the nearest $(R, G, B)$ image value. This step captures the static classes with reasonable fidelity. However, it cannot capture dynamic classes of interest (like \textit{vehicles}) because of the inherent nature of the operation of temporal aggregation of point clouds.
For dynamic classes of interest (like \textit{vehicles}), we paint the vehicle polygons. To get the color for each vehicle polygon, we project the available annotated vehicle cuboids in the camera images. Assign the mode value from a coarse color intensity histogram for each color channel. We bin $8$-Bit colors into $25$ bins of length $10$ each. We show the pipeline in Fig. \ref{fig:datagen}
For occupancy information in BEV, we follow the approach of \cite{lss,fiery} and use the available map information, cuboid annotations for vehicles, sensor and ego-vehicle poses to generate the occupancy map for the scene. 

\section{Conclusion}
This work proposes a method to estimate appearance and occupancy information in a Bird's-eye View (BEV) of the scene, centered on the ego vehicle. 

We propose an architecture that can reason about the appearance and occupancy information in BEV from a set of $N$ monocular images with a $360^\circ$ total FOV and known intrinsic and extrinsic parameters. We train our method on data generated from Carla \cite{carla}. We carry out extensive qualitative and quantitative experiments to capture the efficacy of our system in estimating appearance and occupancy information in different traffic scenarios. We compare our method with SOTA approaches like \cite{lss,neat} and report significant performance improvement. We also carry out ablation studies in different camera configurations(front $180\degree$ and surround $360\degree$). Ablation studies show how appearance can improve occupancy estimation of the scene. We also present a data generation pipeline for real-world datasets like NuScenes\cite{nuscenes} and show representative generated data using the same.

\section{Future work}
\label{futureworkx}
Foremost, we plan to transfer the network to real-world datasets like NuScenes \cite{nuscenes}, KITTI \cite{kitti}, and Lyft \cite{lyft}, and conduct detailed experiments and ablation studies. We also plan to explore the application of color BEV for Multi-Object Tracking (MOT), where we reason about appearance and occupancy in the same output space to associate objects. We also have plans to prospect the potential for using color BEV to develop a language-based navigation system. 
\bibliographystyle{IEEEtran}
\clearpage
\bibliography{references}

\end{document}